# An Algorithm to find Superior Fitness on NK Landscapes under High Complexity: Muddling Through

**Sasanka Sekhar Chanda**

Professor, Strategic Management, *Indian Institute of Management Indore*
Email: sasanka2012@gmail.com

**Sai Yayavaram**

Professor, Strategy, *Indian Institute of Management Bangalore*
Email: sai.yayavaram@iimb.ac.in
*Initial Version:* June 06, 2020. *Updated:* 07 September, 2020

-------------------------------------------------------------------------------

**Abstract.** Under high complexity—given by pervasive interdependence between constituent elements of a decision in an **NK** landscape—our algorithm obtains fitness superior to that reported in extant research. We distribute the decision elements comprising a decision into clusters. When a change in value of a decision element is considered, a forward move is made if the aggregate fitness of the cluster members residing alongside the decision element is higher. The decision configuration with the highest fitness in the path is selected. Increasing the number of clusters obtains even higher fitness. Further, implementing moves comprising of up to two changes in a cluster also obtains higher fitness. Our algorithm obtains superior outcomes by enabling more extensive search, allowing inspection of more distant configurations. We name this algorithm the *muddling through* algorithm, in memory of Charles Lindblom who spotted the efficacy of the process long before sophisticated computer simulations came into being.

-------------------------------------------------------------------------------

*"The most frequent and basic objection …"* (to muddling through in policy-making) *"… is … to the political practice of change only by increment …"* (i.e.) *"to the incremental politics to which incremental analysis is nicely suited."*

−Charles Lindblom (1979: 520)

# Introduction

Research in multiple disciplines—physics [1], evolutionary biology [2], management [3], artificial life [4] to name a few—motivate problems as one of finding high peaks on an **NK** model landscape. In the **NK** model, a decision contains $N$ decision elements or nodes. Each node can take values of either "0" or "1". A *decision configuration* is an instantiation of a decision, with all nodes filled with values of either "0" or "1". The extent of ***Fitness*** of a decision configuration is computed by summing the *fitness contribution* of the $N$ individual nodes making up a decision, and dividing the result by $N$. The *fitness contribution* by an individual node is jointly dependent on its value ("0" or "1") and the values in $K$ other nodes with which the focal node shares a dependency. A matrix having $2^{(K+1)}$ rows and $N$ columns—that we refer to as the *fitness matrix*—is populated with random draws from the *Uniform Distribution*, to enable calculation of the fitness contribution by a node.

A common problem formulated through the **NK** model concerns finding decision configurations with the highest (or lowest) ***Fitness*** values. This is considered an NP-hard problem. This is because, as $N$ increases, the number of feasible decision configurations increases exponentially. For example, if $N = 16$, it is necessary to examine $2^N = 65,536$ decision configurations; for $N = 20$, it is necessary to examine over 1 million configurations.

**Prior research on search for high fitness locations in the NK landscape**

Prior research has suggested several approaches to find high peaks in the **NK** landscape. In a majority of the approaches, search commences from a randomly chosen starting point or *initial decision configuration*, fashioned by populating an $N$-bit (decision) string with values "0" or "1" with equal probability (one-half) and executes for a pre-specified number of time steps.

(I) In the *steepest ascent* (**SA**) approach [**4**], the fitness values of all neighbors differing in value in just one bit (or node) from the current decision configuration (i.e. neighbors at a hamming distance of one unit) are compared with the fitness of the current decision configuration. A move is made to the neighbor having the highest fitness, if the fitness value is higher than fitness at current configuration. The process is repeated till a point is reached where there exist no hamming-one neighbor with a higher fitness or when the clock runs out.

(II) In the *centralized search* (**CS**) approach (also referred to as *local search* [**3**]), a move is made if a randomly-chosen hamming-one neighbor has fitness higher than the fitness at current configuration. As before, search terminates when it is not possible to find a hamming-one neighbor with higher fitness or after execution a pre-specified number of time steps.

(III) Kauffman and colleagues [**1**] describe a *parallel updating* (**PU**) approach to find peaks superior to those obtainable from **CS**. Decision elements (nodes) are accorded a certain probability, τ (0< τ <1) of flipping. In a given generation, all $N$ decision elements attempt to flip in parallel, with probability τ. However, the nodes that are actually allowed to flip are the ones where overall higher fitness is accomplished if solely that node flipped. The process is continued for a pre-specified number of generations, or till a point is reached when it is not possible to obtain higher fitness by flipping one node. The latter stopping criterion is identical to the stopping criterion of the **CS** approach.

**Design of the *muddling through* algorithm**

We call our algorithm the *muddling through* (**MT**) approach. This is in honor of the decision-making process of the same name given by Lindblom [5], in the context of making policy decisions under high complexity. In our algorithm, at initialization, we distribute the set of $N$ decision elements into a number of clusters ($\geq 2$). A decision node is selected at random, for flipping. A move is accepted if the sum of fitness contributions of the co-members in the cluster where the focal node resides is higher than the corresponding value prior to the move; else, a different node is flipped and the calculation is repeated. The walk continues for the permitted number of time steps, or when no hamming-one neighbors can be found where the fitness contribution of the changed cluster exceeds the focal cluster's present (fitness) contribution. The decision configuration having the highest fitness on the way is selected as the outcome of the search process. In effect, reconnaissance of the landscape happens by myopic consideration of fitness contribution of a cluster, but moves are finalized only when a configuration having overall higher fitness is found.

**Results**

**Figure 1.** Comparison of *Fitness* outcomes from alternative search algorithms

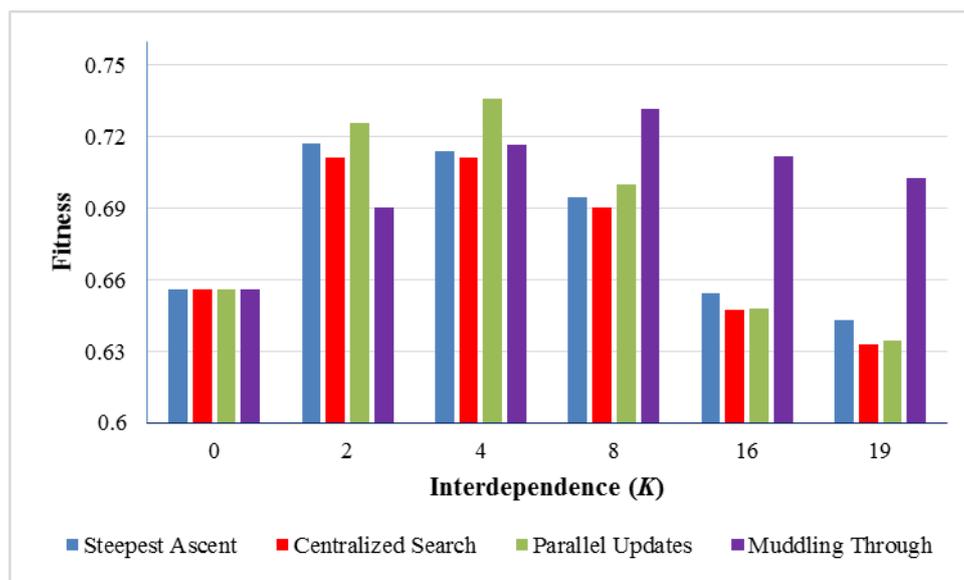

*Note.* For the algorithm "parallel update", the maximum fitness obtained—given varying probability ($\tau$) of nodes updating in parallel—is considered.

In Figures **1**, **2** and **3** we use $N = 20$ and limit search to 1,000 time steps and use four clusters in the *muddling through* algorithm. Results shown are averages over experiments on 10,000 landscapes. In **Figure 1** (above) we demonstrate that in a landscape with $N = 20$, the *muddling through* approach obtains higher fitness outcomes compared to the other alternatives—*steepest ascent*, *centralized search*, *parallel updating*—for $K \geq 8$, i.e. under high complexity.

**Figure 2.** Comparison of *Hamming distance* between initial and final decision configurations, in alternative search algorithms

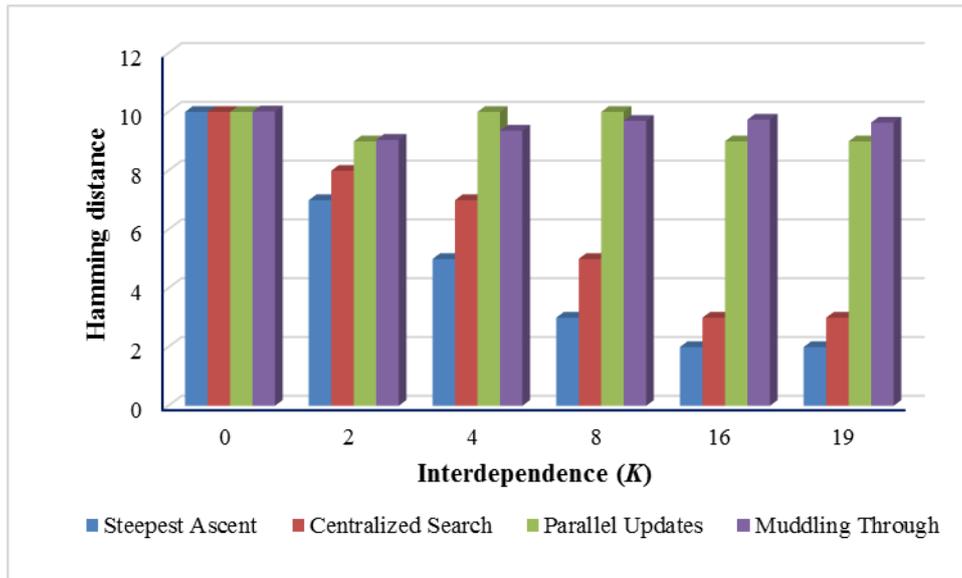

In **Figure 2** we plot the hamming distance between initial and final decision configurations. It shows that *muddling through* retains the ability to explore distant configurations under high $K$.

**Figure 3.** Average resource consumption in alternative search algorithms

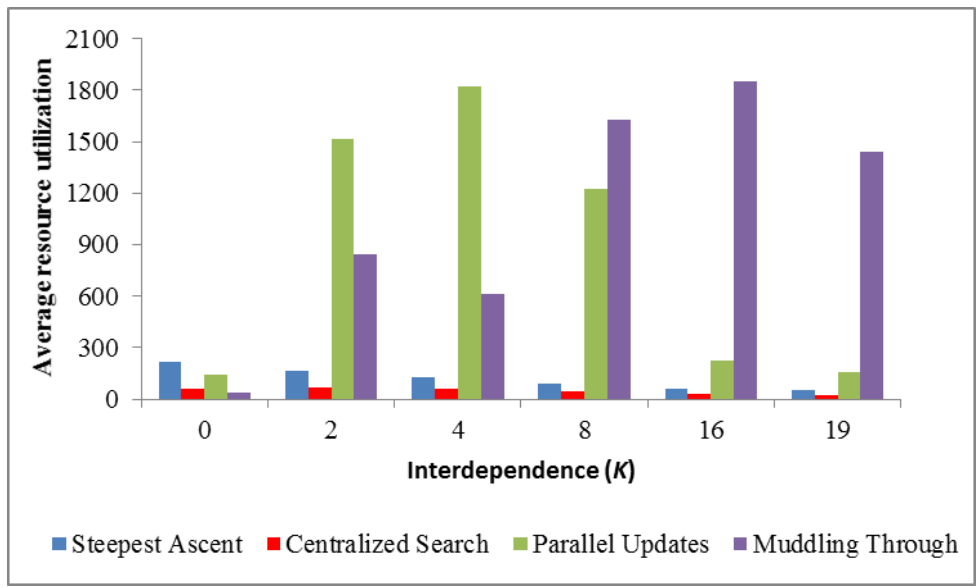

**Figure 3** provides a comparison of the number of moves—flipping of decision nodes to calculate fitness—across the algorithms. We observe that *muddling through* is rather inefficient under low and moderate complexity ($K \leq 4$), given that **SA** and **CS** use much less resources to accomplish comparable or better outcomes. At higher complexity, *muddling through* utilizes resources better, given that other algorithms terminate earlier and attain lower fitness.

## Improving *Muddling Through* to obtain even higher fitness

Under high complexity engendered by pervasive interdependence, we can get higher fitness values by using six clusters instead of four (**Figure 4A**). We show another potent way to obtain higher fitness by the *muddling through* algorithm in **Figure 4B**. Instead of limiting the number of changes to just one (*MT1*) we allow up to two changes per cluster in a given move (*MT2*).

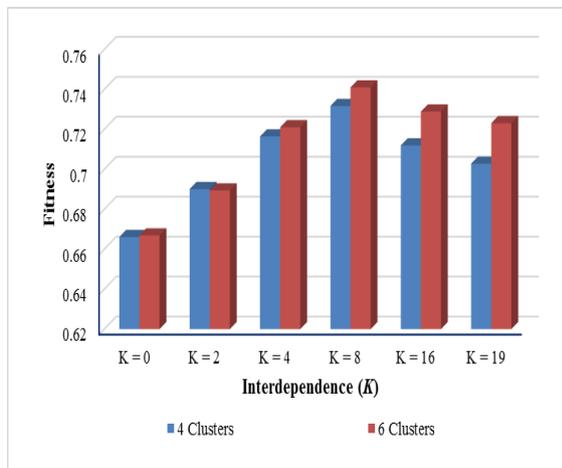

**Figure 4A**. Fitness attained by *Muddling Through* for **4** and **6** clusters

*Parameters*. $N = 20$, Time steps $=1000$, *MT1*. *MT1* signifies *muddling through* with one change per cluster.

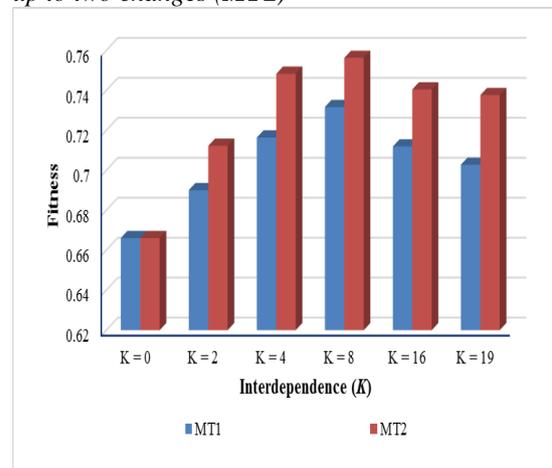

**Figure 4B**. Fitness attained by *Muddling Through* with one change (*MT1*) vs Muddling Through with up to two changes (*MT2*)

*Parameters*. $N = 20$, Time steps $=1000$, 4 Clusters. *MT1* signifies *muddling through* with one change per cluster. *MT2* signifies *muddling through* with up to two changes per cluster.

## Discussion

In low and moderate complexity ($K \leq 3$), where there are fewer peaks, the discovery of a high fitness peak may be thwarted altogether, under search by *muddling through*. The *muddling through* algorithm is blind to a subset of approach paths to fitness peaks. Non-accessing of a subset pathways to scarce fitness peaks results in attainment of lower fitness configuration in low and moderate complexity ($K \leq 3$). When $K > 3$ the fitness landscape is increasingly uncorrelated [6] but peaks are more numerous. Thus, some highly advantageous peaks exist, that cannot be reached by moving only to successively higher fitness points from the beginning.

Since search by *muddling through* permits getting on to lower fitness locations, it is able to access paths to some advantageous peaks that are denied to search by other approaches.

The progress reported in this paper has applications in a number of fields. Limiting ourselves to the field of policy decisions, we observe that beneficial radical change—or valuable far-reaching adaptation—is more likely to materialize by the approach of *muddling through*. This later finding confounds conventional wisdom in the policy field [**7**], but is in line with Lindblom's claims [**8**] regarding the potency of his approach—as outlined in our opening quotation.